\documentclass[twoside,leqno,twocolumn]{article}

\usepackage[letterpaper]{geometry}
\usepackage[font=small]{caption}
\usepackage{ltexpprt}
\usepackage{subcaption}
\usepackage{hyperref}
\usepackage{url}
\usepackage{graphicx}
\usepackage{wrapfig}
\usepackage{cite}
\usepackage{amsmath}
\usepackage{amssymb}
\usepackage{amsfonts}
\usepackage{bm}
\usepackage{physics}
\usepackage{placeins}
\usepackage{xcolor}
\usepackage{float}
\usepackage{flafter}
\begin{document}
 
\title{Quadratic Residual Networks: A New Class of Neural Networks for Solving Forward and Inverse Problems in Physics Involving PDEs}
\author{Jie Bu\footnotemark[1]
\and Anuj Karpatne\footnotemark[1]}

\date{}

\maketitle

\footnotetext[1]{Discovery Analytics Center, Dept.\ of Computer Science, Virginia Tech, VA, USA}

\makeatletter
\newcommand{\thickhline}{%
    \noalign {\ifnum 0=`}\fi \hrule height 0.65pt
    \futurelet \reserved@a \@xhline
}

\newcommand{\fix}{\marginpar{FIX}}
\newcommand{\new}{\marginpar{NEW}}
\newcommand{\sym}{\textup{Sym}}
\renewcommand{\dim}{\textup{dim}\ }
\makeatother


\fancyfoot[R]{\scriptsize{Copyright \textcopyright\ 2021 by SIAM\\
Unauthorized reproduction of this article is prohibited}}

\begin{abstract} \small\baselineskip=9pt 
We propose \textit{quadratic residual networks} (QRes) as a new type of parameter-efficient neural network architecture, by adding a quadratic residual term to the weighted sum of inputs before applying activation functions. With sufficiently high functional capacity (or expressive power), we show that it is especially powerful for solving forward and inverse physics problems involving partial differential equations (PDEs). Using tools from algebraic geometry, we theoretically demonstrate that, in contrast to plain neural networks, QRes shows better parameter efficiency in terms of network width and depth thanks to higher non-linearity in every neuron. Finally, we empirically show that QRes shows faster convergence speed in terms of number of training epochs  especially in learning complex patterns.
\end{abstract}

\section{Introduction}

Deep neural networks (DNNs) have found remarkable success in a number of learning tasks, thanks to their ability to approximate arbitrarily complex functions \cite{kidger2020universal,generalization1989}. 
The composition of many nonlinearly activated neurons gives DNN high functional capacity despite the weighted sum being linear in every neuron.
Intuitively, a DNN with higher capacity has a better ability to capture complex patterns from data in lesser number of training epochs. However, in order to learn generalizable patterns, we often need to proportionately balance the capacity of a DNN with the \textit{amount of supervision} available in the training data. Indeed, it is common practice to use regularization tools such as dropout and early stopping to avoid \textit{overfitting} the DNN model to complex spurious patterns in the training set, especially when training sizes are very small.

\begin{figure}[ht]
    \centering
    \includegraphics[width=0.42\textwidth]{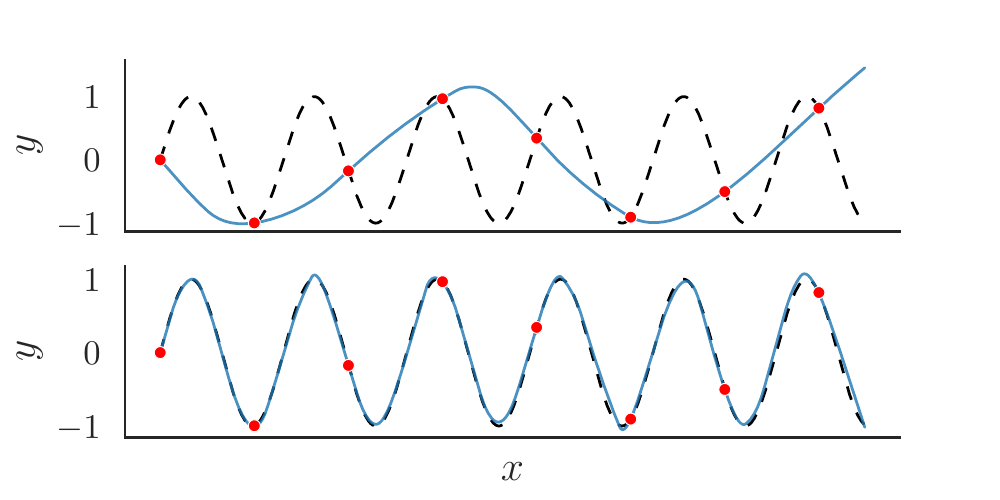}
    \vspace{-2ex}
    \caption{Toy demonstration of the importance of physics supervision in learning a target function $y=\sin(6x)$ (dashed) given limited data (8 red points). Top and bottom figures show neural network solutions (solid) trained solely using data, and both data and physics (satisfying the simple PDE constraint: $\frac{dy}{dx} = 6\cos(6x)$), respectively.}
    \label{fig:pde_sin}
    \vspace{-4ex}
\end{figure}

While data-driven supervision is limited in conventional learning tasks, there is a growing body of research on using {physical knowledge} as another form of supervision to train machine learning (ML) models, termed as the field of \textit{physics-guided machine learning} (PGML) \cite{karpatne2017theory,willard2020integrating}. In this work, we specifically study the class of problems in PGML where the physics of the system is available in the form of partial differential equations (PDEs).
A promising line of work in this area is the framework of  \textit{physics-informed neural networks} (PINNs) \cite{raissi2017physics1,raissi2017physics2}, where a neural network is used to model a target variable $u$ (e.g., velocity field) given some inputs (e.g., location $x$ and time $t$), based on the \textit{physical constraint} that $u(x,t)$ satisfies a known PDE. In PINNs, the neural networks are trained not only using supervision from data (by minimizing prediction errors on labeled points) but also from physics (by evaluating the consistency of neural network predictions with PDE on plentiful unlabeled points).

\begin{figure}[t]
    \begin{center}
        \begin{subfigure}[b]{0.40\textwidth}
            \centering
            \includegraphics[width=0.9\textwidth]{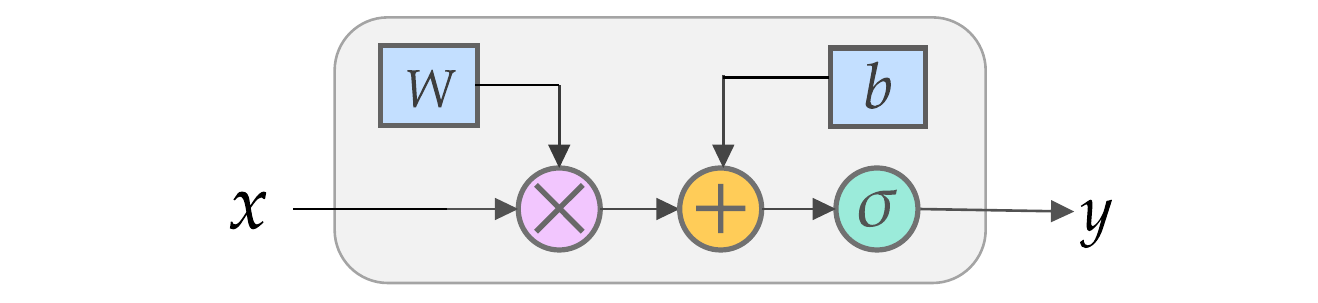}
            \caption{Plain DNN Layer}
            \label{fig:NN}
        \end{subfigure}
        \\[2ex]
        \begin{subfigure}[b]{0.40\textwidth}
            \centering
            \includegraphics[width=0.9\textwidth]{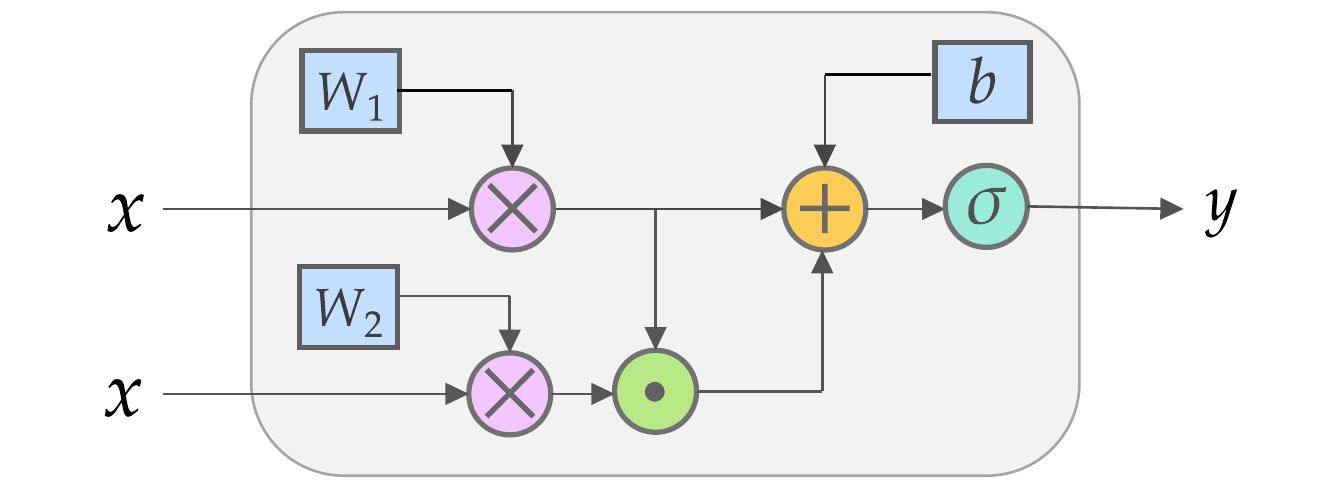}
            \caption{Quadratic Residual Network (QRes) Layer}
            \label{fig:QRes}
        \end{subfigure}
    \end{center}
    \vspace{-2ex}
    \caption{Overview of our proposed Quadratic Residual Network (QRes) layer in comparison with plain DNN layer. Blue rectangular boxes represent trainable parameters and round boxes represent operations (purple ``$\times$": multiplication, orange ``$+$": addition, green "$\cdot$": Hadamard product, and cyan "$\sigma$": activation operator).}
    \label{fig:layer_structure}
    \vspace{-3.5ex}
\end{figure}

The additional supervision from physics enables PINN to support neural networks with sufficiently high capacity 
without running into risks of overfitting (see Figure \ref{fig:pde_sin} for a demonstration on a toy problem). This has sparked ample interest in the scientific community to use PINNs and their variants in a number of physical problems involving PDEs \cite{Raissi1026,jin2020nsfnets,gao2020phygeonet}. Despite these developments, most existing work in PINN only uses plain DNN architectures (see Figure \ref{fig:NN}).
As a result, PINN formulations typically require a vast number of network parameters and training epochs to approximate complex PDEs with acceptable accuracy \cite{raissi2019physics}. This motivates us to ask the question: \textit{Can we develop a neural network architecture with higher capacity at every layer that can approximate complex functions with less parameters than plain DNNs?}

We present \textit{Quadratic Residual networks} (QRes), a novel class of neural network architectures that impart quadratic non-linearity before applying activation functions at every layer of the network. Figure \ref{fig:QRes} shows an overview of a QRes layer where a quadratic residual term: $W_1x \circ W_2x$ is added to the weighted sum $W_1x + b$ of a plain DNN layer before passing through a non-linear activation $\sigma$.
We theoretically study the expressive power of QRes to demonstrate that QRes is more parameter efficient than plain DNNs. We also conduct extensive experiments on forward and inverse problems involving PDEs by replacing DNNs with QRes in PINN frameworks and demonstrate better parameter efficiency of QRes over baselines. Finally, we empirically show that QRes converge faster than plain DNNs especially in learning higher frequency patterns.

\section{Background}
\label{sec:background}


\subsection{Physics-informed Neural Networks:}

\begin{figure}[t]
    \centering
    \includegraphics[width=0.49\textwidth]{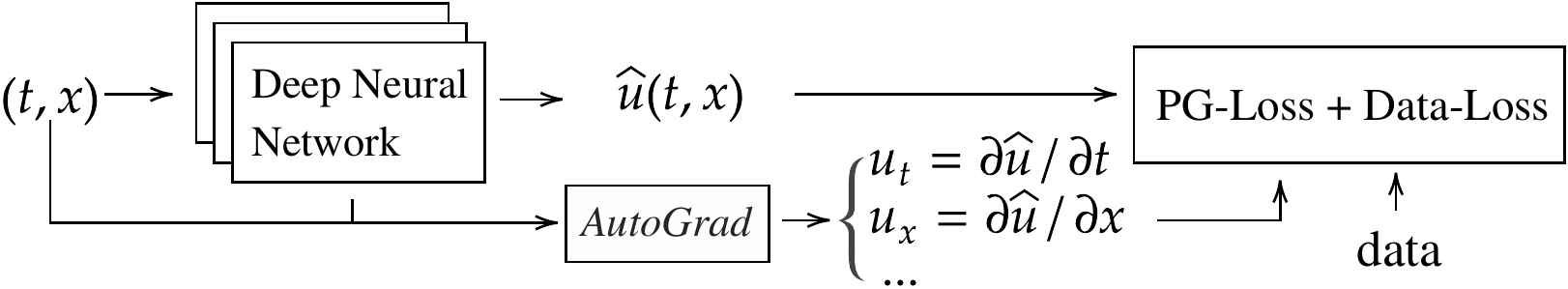}
    \vspace{-2ex}
    
    \caption{The general framework of PINN. 
    }
    \label{fig:pinn}
    \vspace{-4ex}
\end{figure}

There is growing body of work in the field of PINN for guiding the learning of neural networks using physics supervision available as PDEs 
\cite{raissi2019physics,raissi2017physics1,raissi2017physics2,Raissi1026,jin2020nsfnets,gao2020phygeonet}. A general form of a non-linear PDE can be expressed as $\mathcal{N}(u,\lambda) = 0$, where $\mathcal{N}$ is a non-linear operator involving partial derivatives of the target variable $u$ (e.g., $u_x, u_t, u_{xx}, \ldots$), and $\lambda$ represents the parameters of the PDE. There are two classes of problems in the realm of PDEs that are studied by PINNs: (a) \textit{forward problems}, where the goal of the  network is to solve for the target variable $u$ satisfying the PDE, and (b) \textit{inverse problems}, where the  network is tasked to learn the unknown parameters of the PDE, $\lambda$, given ground-truth values of $u$ at a collection of points.

In both classes of problems, a central form of supervision used for training neural networks is the discrepancy of the network predictions w.r.t. the governing PDEs, captured as physics-guided $\text{(PG)-Loss} = \mathcal{N}(\hat{u},\lambda)^2$. This loss is evaluated at a set of unlabeled points, where the partial derivatives involved in PG-Loss are computed analytically using automatic differentiation tools (e.g., AutoGrad). 
The neural networks are also supervised with data-driven loss over a set of labeled points $S$, $\text{Data-Loss} = \sum_{i\in S}||\hat{u}_i - u_i||^2$, where $S$ usually comprises of initial or boundary condition points. The combined learning objective in PINN is thus to minimize PG-Loss $+$ Data-Loss (see Figure \ref{fig:pinn}).

To accelerate training convergence of PINNs, adaptive activation functions have recently been proposed in \cite{Jagtap_2020}, where a learnable scalar $\alpha$ is multiplied to the weighted sum ($Wx$) produced at every layer before applying activation. There has also been recent studies exposing gradient pathologies in the training of PINNs \cite{wang2020pinns}, where adaptive trade-off between loss terms were proposed to resolve the imbalance in loss gradients. In the same work \cite{wang2020pinns}, the authors further explored a modified neural network architecture for PINN problems, where, inspired by attention mechanisms, they explicitly accounted for multiplicative terms involving inputs in the network layers, similar to our work. While it empirically showed the importance of using higher-order interactions in PINN frameworks, it did not provide any theoretical justifications for its effect on the expressive power of neural networks in a principled manner as performed in our work.




\subsection{Related Work on Quadratic Networks:}
There is a long history of research on building neural networks to capture multiplicative interactions among inputs, ranging from early works on optimal depths of plain DNNs for approximating multiplications \cite{siu1993optimal} to the use of weighted products instead of summations at every unit of the network \cite{durbin1989product,urban2015neural}. In the area of graphical models, the framework of \textit{sum-product networks} (SPNs)  \cite{poon2012sumproduct} have been developed to represent factorization operations (product nodes) in addition to mixture operations (sum nodes) for learning partition functions. Our work shares a similar motivation as SPNs to learn more expressive functions for approximating complex decision boundaries. Our work is also related to the recent framework of \textit{neural arithmetic units} (NAUs) \cite{trask2018neural,madsen2020neural}, that perform exact arithmetic operations (e.g., additions, subtractions, and multiplications) at every layer to logically extrapolate well on arithmetic tasks. Our work can be viewed as a special case of NAUs that capture quadratic residual terms, although for a different goal of expressing higher functional capacity in the process of solving non-linear PDEs.

Another line of work that bears close resemblance to our work is the quadratic deep networks (QDNs) \cite{fan2018new}, where three weight matrices are used to express quadratic products as well as linear sums at every layer before applying activations. Our work is different from QDNs on two grounds. First, we provide novel theoretical analyses of the expressive power of QRes that proves its superior parameter efficiency over plain DNNs. Second, in contrast to QDNs, we demonstrate the efficacy of using QRes in solving PINN problems, where neural networks with higher functional capacity can be better supported with the aid of physics supervision, in contrast  to conventional learning tasks that only use data-driven supervision.




\section{Quadratic Residual Networks}
\label{sec:method}

A plain DNN layer can be expressed as $y^{DNN} = \sigma(Wx + b)$, where $(W,b)$ are the learnable parameters and $\sigma$ is a non-linear activation function. Notice that $Wx + b$ is linear and it is only $\sigma$ that imparts non-linearity to the outputs. As a result, we need a large number of DNN layers with reasonable widths to capture sufficient non-linearity with acceptable accuracy.

In contrast, we consider \textit{quadratic residual terms} at every layer of our QRes network to contribute additional nonlinearity. In particular, we can express a single layer of QRes as $y^{QRes} = \sigma({\color{red}{W_2x\circ W_1x}} + W_1x + b)$, where $\circ$ denotes the Hadamard product and the term in {\color{red} red} is the quadratic residual term (we call it ``residual'' as removing it simply yields a plain DNN). Hence, in problems where linear decision boundaries (activated non-linearly) are sufficient to capture the complexity of target functions, QRes can easily resort to a plain DNN by learning $W_2 = 0$. However, in problems where we need neural networks with higher functional capacity than DNNs, QRes can switch on the quadratic residual term to capture higher amounts of non-linearity using efficient network depths and widths.

It is easy to show that a linearly activated QRes (using linear activations) with depth $d$ can learn polynomials of degree $2^{d-1}$, since every layer of QRes would double the non-linearity by considering products of outputs from previous layer. As a result, even a linearly activated QRes, in theory, can approximate arbitrarily complex polynomial boundaries with sufficient network widths and depths. However, in practice, it is desirable to use non-linearly activated QRes for two reasons. First, non-linearly activated QRes can approximate polynomial decision boundaries using smaller network depths than a linearly activated QRes, thus resulting in parameter efficiency. Second, a linearly activated QRes with a large number of layers can produce unbounded activation outputs at every layer, which, if not properly scaled, can lead to unstability in training. Hence, we use non-linear activations with bounded output spaces ($tanh$) in all our implementations of QRes.




\section{Theoretical Analyses of QRes}

To analyze the expressive power of QRes on regression tasks, we draw inspiration from theoretical analyses of the expressive power of \textit{deep polynomial networks} (networks with polynomial activation functions) presented in \cite{kileel2019expressive}, using concepts from \textit{algebraic geometry}. 

\subsection{Definitions and Notations:} 

Let us represent a network architecture\footnote{Note that the \textit{architecture} $\bm{d}$ in this section refers to a sequence of layer widths
.} as a vector, $\bm{d} = \{d_0, \ldots,d_h\}$, where $d_i$ is the width of layer $i$ and $h$ is the network depth. To understand the space of functions expressed by such a network architecture, let us consider the functional mapping from the space of network parameters to the space of output functions. The image (set of outputs) of this functional mapping is referred to as the \textit{functional space} associated with the network. The dimension of the functional space gives us a measure of the expressive power of a network.

In this work, we characterize the functional space of a network using a basis set of polynomials (obtained, for example, using Taylor approximations). We also consider a special type of neural networks with polynomial activation functions of fixed degree $r$, which raises the input to their $r$-th power, namely \textit{polynomial networks}. As shown in \cite{kileel2019expressive}, the functional space of a polynomial network comprises of \textit{homogeneous polynomials}, i.e., polynomials where every term (monomial) is of the same degree. We denote the space of all homogeneous polynomials of degree $d$ in $n$ variables with coefficients in $\mathbb{R}$ as $\sym_{d}{(\mathbb{R}^{n})}$. 

Under these settings, the family of all networks with the same architecture $\bm{d}$ can be identified with its \textit{functional variety}, which is the \textit{Zariski closure}\footnote{The Zariski closure of a set $\mathcal{X}$ is the smallest set that contains $\mathcal{X}$ and can be described by polynomial equations \cite{kileel2019expressive}.} of its functional space. An advantage of analyzing functional variety with polynomial equations is that it requires less strict assumptions compared to the functional space \cite{kileel2019expressive} and has the same dimension as the functional space, making the dimension of the functional variety of a network a precise measure of the networks' expressiveness.

Note that even with polynomial activation functions, the functional space of a QRes network is not a space of homogeneous polynomials. For example, for a single layer QRes network with linear activations, its functional space contains spaces of homogeneous polynomials of degrees both 1 and 2. Furthermore, neural networks with widely-used activation functions \cite{datta2020survey} also result in non-homogeneous functional spaces because popular non-linear activation functions (e.g., \textit{tanh}) are generally not homogeneous polynomial mappings. However, every space of non-homogeneous polynomials can be viewed as comprising of subspaces of homogeneous polynomials of varying degrees. Thus, to generalize the analysis of functional spaces beyond polynomial networks, we introduce the definition of the \textit{leading functional space} of a network as the subspace of its functional space comprising of homogeneous polynomials of highest degree. In general, for any activation function, we can always decompose it to a set of polynomial functions using Taylor approximations. The highest degree of polynomial, $r$, in such a decomposition can then be referred to as the \textit{leading degree} of the activation function. 

Formally, for a network architecture $\bm{d}$ with an activation function of leading degree $r$, we denote the \textit{leading functional space} of a neural network as $\mathcal{F}_{\bm{d}, r}$ and a QRes network as $\mathcal{F}^{2}_{\bm{d}, r}$. The leading functional variety of a neural network and a QRes network can then be defined as the Zariski closure of its leading functional space, i.e., $\mathcal{V}_{\bm{d}, r} = \overline{\mathcal{F}^{2}_{\bm{d}, r}}, ~\text{and}~\mathcal{V}_{\bm{d}, r} = \overline{\mathcal{F}^{2}_{\bm{d}, r}}$.
Using these definitions, we introduce the revised concepts of \textit{filling} functional space and variety (similar to the ones presented in \cite{kileel2019expressive}) as follows:

\begin{Definition}
A neural network architecture $\bm{d}=(d_0, ..., d_h)$ has a filling functional space for the activation degree $r$ if its leading functional space satisfies $F_{\bm{d},r} = \sym_{r^{h-1}}(\mathbb{R}^{d_0})^{d_h}$. For a filling functional variety, its leading functional variety satisfies $V_{\bm{d},r} = \sym_{r^{h-1}}(\mathbb{R}^{d_0})^{d_h}$.
\label{def:leading}
\end{Definition}

Hence, rather than requiring the functional space or variety of a network to fill the ambient space of homogeneous polynomials, we only require it to \textit{contain} the space of homogeneous polynomials of leading degree for it to be considered as filling. 

\subsection{Theoretical Analyses:} 
Proofs of all theoretical analyses introduced in this work are available in the Appendix section \ref{sec:proofs}.

\begin{proposition}
\label{pp:single_layer_QRes}
A single-layer linearly activated \textup{($r = 1$)} quadratic residual networks of architecture $\bm{d}=(d_0, d_1)$ has a filling functional space of degree 2, i.e., its leading functional space $\mathcal{F}^{2}_{\bm{d}, 1} = \sym_{2}{(\mathbb{R}^{d_0})^{d_1}}$.
\end{proposition}

A noticeable feature of linear neural networks is that the degree of its functional variety will not grow with the network depth while the degree of the functional space of a linear QRes network can grow exponentially with network depth. The growing degree of QRes suggests it can obtain more nonlinearity from deep architectures. A QRes network with linear activation can be related to a polynomial regression, where a single layer corresponds to a quadratic regression. Proposition \ref{pp:single_layer_QRes} can be easily generalized to QRes with deep architectures as follows. 

\begin{lemma} 
\label{lm:efficiency}
For an activation function with leading degree $r \geq 1$ and network architecture $\bm{d} = (d_0, ..., d_h)$, the leading functional variety of a QRes network, $\mathcal{V}^{2}_{\bm{d}, r}$, and a neural network, $\mathcal{V}_{\bm{d}, r}$, satisfy $\mathcal{V}^{2}_{\bm{d}, r} = \mathcal{V}_{\bm{d}, 2r}$.
\end{lemma}

The above lemma states that, with the same network architecture, a QRes network with leading activation degree $r$ and a neural network with leading activation degree $2r$ have functional varieties of the same degree of homogeneous polynomials. This implies that a deep QRes network can have a leading functional variety of degree $(2r)^{h-1}$ homogeneous polynomials, while a neural network of the same architecture and activation function can only reach a degree of $r^{h-1}$. Note that this lemma does not require the functional variety of either networks to be filling, and it holds both for linear and nonlinear activations. Using this property we can arrive at the following theorem.

\begin{theorem}
\label{th:depth_efficiency}
\textup{(Depth Efficiency)} For a fixed leading degree $r \geq 2$, let us assume that the functional variety of a quadratic residual network $\bm{d_q} = (d_0, .., d_{h_q})$ is filling, $\mathcal{V}^{2}_{\bm{d_q}, r}$ is the leading functional variety for the QRes network, and $\mathcal{V}_{\bm{d}, r}$ is the leading functional variety for a neural network $\bm{d_n} = (d_0', .., d_{h_n}')$, where $h_n, h_q > 1$ and $d_0 = d_0', d_{h_q} = d_{h_n}'$. If $\dim{\mathcal{V}_{\bm{d_n}, r}} \geq \dim{\mathcal{V}^{2}_{\bm{d_q}, r}}$, then
\begin{equation}
h_n \geq 1 + \left( 1 + \frac{\log{2}}{\log{r}} \right)(h_q - 1)    
\end{equation}
\end{theorem}

The above theorem throws light on the depth efficiency of QRes as it provides a lower bound on the depth of a neural network $h_n$ for it to have greater expressibility (i.e., larger dimension of functional variety) than a filling QRes network. Although $h_n$ and $h_q$ will converge for large values of $r$, it may need an extremely wide network to be able to have a filling functional variety of high degrees. Therefore, it is necessary to show the efficiency of QRes in terms of network width, presented in the following.

\begin{proposition}
\label{pp:bound_filling_with_qres}
\textup{(Minimal Filling Width)} For a neural network or a QRes network with architecture $\bm{d} = (d_0, ..., d_h)$ and leading activation degree $r \geq 2$, if 
\begin{equation}
d_{h-i} \geq \min{\left[ d_h r^{id_0}, \binom{r^{h-i} + d_0 - 1}{ r^{h-i} } \right]}    
\end{equation}
for each $i = 1, ..., h-1$, then its functional variety is filling, and we call the lower bound of $d_{h-i}$ as the minimal filling width of this layer with leading degree $r$.
\end{proposition}

We refer to the architecture with minimal filling width at each intermediate layer as the \textit{minimal filling architecture}. Using \textit{Proposition \ref{pp:bound_filling_with_qres}}, we can arrive at the following theorem for width efficiency.

\begin{theorem}
\label{th:width_efficiency}
\textup{(Width Efficiency)} Suppose a neural network $\bm{d_n} = (d_0, .., d_{h})$ is filling for leading activation degree $r \geq 2$. Given a quadratic residual network $\bm{d_q} = (d_0', .., d_{h}')$ with $d_0 = d_0'$ and  $d_{h} = d_{h}'$, such that $\dim{\mathcal{V}_{\bm{d_n}, r}} = \dim{\mathcal{V}^{2}_{\bm{d_q}, r}}$. Suppose $\bm{d_n}$ is a minimal filling architecture, then for each $i = 1, ..., h-1$,  
\begin{equation}
\lim_{r \to \infty}{d_{h-i}} = O(2^{\tau}) \lim_{r \to \infty}{d_{h-i}'}    
\end{equation}
where $\tau = \min{\left[ id_0, (h-i)(d_0-1) \right]}$.
\end{theorem}

The above theorem shows that a QRes network is exponentially more efficient than a neural network in terms of width to achieve the same expressive power (i.e., dimension of functional variety). Since the number of network parameters grow roughly linearly with network depth but polynomially with the network width, width efficiency is a dominating factor in the overall parameter efficiency.
Further, while the above analysis was performed using polynomials as the basis set, it is easy to extend this analysis to frequencies in the spectral space (by applying Fourier decomposition). Since QRes can express higher degree of polynomials more efficiently than neural networks, QRes is also able to capture higher frequency information with comparable or even smaller number of parameters, as shown empirically in Section \ref{sec:eval_patterns}. 



\begin{figure}[tb]
    \begin{center}
    \includegraphics[width=0.49\textwidth]{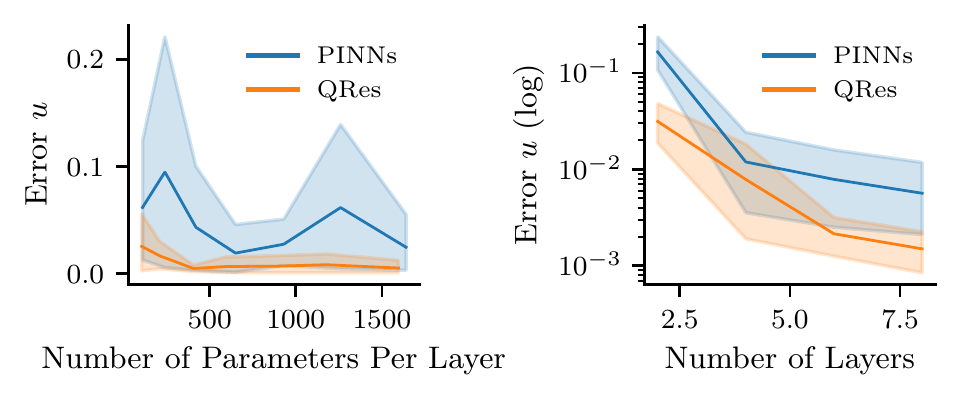}
    \vspace{-3ex}
    \caption{Parameter efficiency of PINNs and QRes networks for solving Burgers' equation.  The figures show performances across 28 architectures over four values of depth (2, 4, 6, 8), and seven widths: (10, 15, 20, 15, 30, 35, 40) for PINNs and (7, 10, 14, 17, 21, 24, 28) for QRes. Lines indicate mean values and shades represent variances.}
    \label{fig:burgers_efficiency}
    \end{center}
    \vspace{-3ex}
\end{figure}

\section{Empirical Results}
\label{sec:results}

\begin{table*}[ht]
\begin{center}
\begin{small}
\caption{\normalsize Set of forward and inverse PDE problems studied in this work.}
\begin{tabular}{lll|l}
\thickhline 
PDE                              & Problem Type   & Model Type      & Descriptions                                                                                                                                                                            \\ \hline
Navier-Stokes (N-S)                     & Inverse & Continuous Time & $\begin{array}{l} u_t + \lambda_1 (u u_x + v u_y) = -p_x + \lambda_2(u_{xx} + u_{yy}),\\ v_t + \lambda_1 (u v_x + v v_y) = -p_y + \lambda_2(v_{xx} + v_{yy}), \\ u_x + v_y = 0, \\ \textup{Given data}\ u, v,\ \textup{predict}\ \lambda_1, \lambda_2, p. \end{array}$              \\ \hline
Burgers'                          & Inverse & Continuous Time & $\begin{array}{l} u_t + \lambda_1 u u_x - \lambda_2 u_{xx} = 0, \\ \textup{Given data}\ u,\ \textup{predict}\ \lambda_1, \lambda_2. \end{array}$                                                                                                                                          \\ \hline
Korteweg–de Vries                & Inverse & Discrete Time   & $\begin{array}{l} u_t + \lambda_1 u u_x + \lambda_2 u_{xxx} = 0, \\ \textup{Given data}\ u,\ \textup{predict}\ \lambda_1, \lambda_2.\end{array}$                                                                                                                                         \\ \hline
Burgers'                          & Forward      & Continuous Time & $\begin{array}{l} u_t + u u_x - (0.01/\pi) u_{xx} = 0, \ x \in [-1,1], \ t \in [0,1],\\ u(0,x) = -\sin(\pi x),\\ u(t,-1) = u(t,1) = 0. \\ \textup{Predict}\ u.\end{array}$                               \\ \hline
Schr\"{o}dinger & Forward      & Continuous Time & $\begin{array}{l} i h_t + 0.5 h_{xx} + |h|^2 h = 0, \ x \in [-5, 5], \ t \in [0, \pi/2],\\ h(0,x) = 2\ \text{sech}(x),\\ h(t,-5) = h(t, 5),\\ h_x(t,-5) = h_x(t, 5), \\ \textup{Predict}\ h.\end{array}$ \\ \hline
Allen-Cahn                       & Forward      & Discrete Time   & $\begin{array}{l} u_t - 0.0001 u_{xx} + 5 u^3 - 5 u = 0,  \ x \in [-1,1],  \ t \in [0,1],\\ u(0, x) = x^2 \cos(\pi x),\\ u(t,-1) = u(t,1),\\ u_x(t,-1) = u_x(t,1). \\ \textup{Predict}\ u.\end{array}$  \\
\thickhline  
\end{tabular}
\label{tab:problems}
\end{small}
\vspace{-3ex}
\end{center}
\end{table*}

We evaluated PINN (using DNN) and QRes on a set of forward and inverse problems (see Table \ref{tab:problems}) involving nonlinear partial  differential equations, same as those used in \cite{RAISSI2019686}.
We include both \textit{continuous time} and \textit{discrete time} models based on definitions from \cite{raissi2017physics1, raissi2017physics2}. Except for the forward problem on Burgers' equation, all models are trained with Adam optimizer for a fixed number of epochs and subsequently trained on L-BFGS-B for better accuracy, following the practices used in previous work on PINNs \cite{raissi2017physics1,raissi2017physics2}. In addition to prediction errors, we also report the number of network parameters (including the bias term) and training epochs on Adam. We observed that the number of epochs needed for L-BFGS-B to reach termination condition is roughly the same for PINNs and QRes on all the experiments, while the convergence speed on Adam plays a dominating role for training efficiency. Additional specifications of experiments are available in the  Appendix section \ref{sec:appendix}, and all codes are on Github\footnote{\url{https://github.com/jayroxis/qres}.}. 

\paragraph{Better Accuracy:}
Tables \ref{tab:identification} and \ref{tab:inference} show the results of the overall evaluation of QRes and PINN on different PDEs. With the same number of parameters and epochs, QRes consistently outperforms PINN, e.g., for Navier-Stokes (N-S) in Table \ref{tab:identification}. Results on Burgers' equation in Table \ref{tab:identification} show that even with less number of parameters, QRes still manages to have better accuracy than PINNs over most of the predictions, which is verified by results from Table \ref{tab:inference}. To push the limit of the QRes networks even further, we reduce the number of network parameters as well as training epochs for the kDV equation in Table \ref{tab:identification}. We can see that QRes maintains better accuracy over PINN with less than half of the PINN's network parameters and 1/5 training epochs.

\begin{table*}[ht]
\caption{Parameter identification on Navier-Stoke (N-S) equation, Burgers' equation and Korteweg–de Vries (KdV) equation. $e(\lambda)$ and $e_n(\lambda)$ represents the percentage errors of predicted parameter of the PDEs $\lambda$ w.r.t. its ground truth values, while $e$ and $e_n$ correspond to experiments on clean data and data with 1\% noise, respectively.}
\begin{center}
\begin{small}
\begin{tabular}{l|l|l|l|l|l|l|l}
\thickhline
PDE     & Model     & Param. & Epochs & $e(\lambda_1)$\% & $e(\lambda_2)$\% & $e_n(\lambda_1)$\% & $e_n(\lambda_2)$\% \\ \hline
N-S     & PINN      & 3.06k  & 200k & 0.083 & 5.834 & 0.077 & 5.482 \\ 
N-S     & QRes      & 3.00k  & 200k & 0.043 & 4.281 & 0.050 & 4.942 \\ \hline
Burgers' & PINN      & 3.02k  & 5k  & 0.057 & 0.636 & 0.170 & 0.031 \\ 
Burgers' & QRes      & 1.54k  & 5k  & 0.027 & 0.379 & 0.172 & 0.003 \\ \hline
KdV     & PINN      & 10.35k & 50k  & 0.017 & 0.011 & 0.154 & 0.045 \\ 
KdV     & QRes      & 4.61k  & 10k  & 0.009 & 0.009 & 0.183 & 0.009 \\ \thickhline
\end{tabular}
\label{tab:identification}
\end{small}
\vspace{-3ex}
\end{center}
\end{table*}



\begin{table*}[ht]
\begin{small}
\caption{Solving Schr\"{o}dinger and Allen-Cahn equations. Results are reported using normalized (by ground truth values) mean square errors between predictions and ground truth values. $u$ and $v$ are the real and imaginary parts of the complex solution $h$ in Schr\"{o}dinger equation, and $|h| = \sqrt{u^2 + v^2}$ is the modulus.}
\begin{center}
\begin{tabular}{l|l|l|l|l|l|l}
\thickhline
PDE             & Model & Param.  & Epochs & Error $u$ & Error $v$ & Error $|h|$       \\ \hline
Schr\"{o}dinger & PINN  & 30.80k  & 50k & 1.456e-03 & 1.878e-03 & 1.099e-03 \\ 
Schr\"{o}dinger & QRes  & 15.60k  & 50k & 1.379e-03 & 1.751e-03 & 1.059e-03 \\ \hline
Allen-Cahn      & PINN  & 141.30k & 10k & 5.044e-03 & N/A       & N/A       \\ 
Allen-Cahn      & QRes  & 25.50k  & 10k & 3.577e-03 & N/A       & N/A       \\ \thickhline
\end{tabular}
\label{tab:inference}
\end{center}
\vspace{-1ex}
\end{small}
\end{table*}




\paragraph{Parameter Efficiency:}
 To further explore the parameter efficiency of PINN and QRes, we experimented with different network widths and depths of both networks for solving Burgers' equation. 
Figure \ref{fig:burgers_efficiency} 
shows how the prediction errors vary with different network widths and depths. 
We can see that the QRes networks outperform PINNs not only under the same settings but even with much less number of parameters. It demonstrates a huge advantage of the QRes networks in terms of parameter efficiency, which strongly supports our theoretical results.


\begin{figure}[ht]
    \centering
    \includegraphics[width=0.49\textwidth]{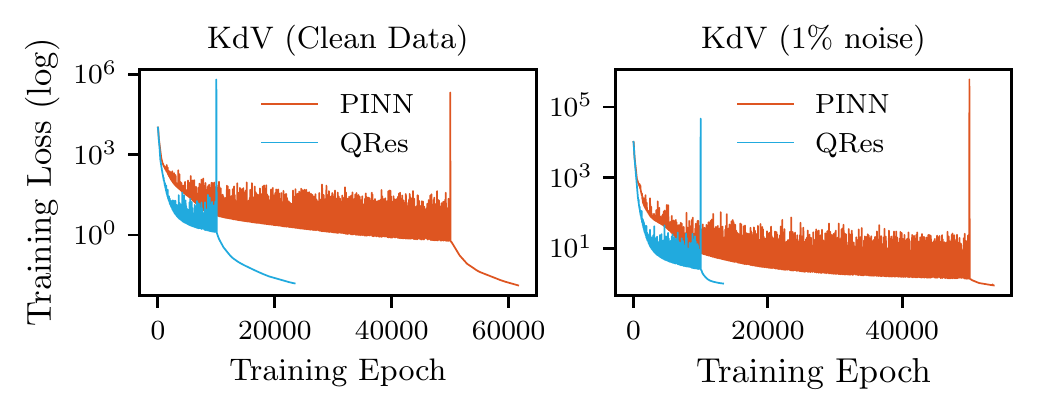}
    \vspace{-3ex}
    \caption{Training loss curves for parameter identifications on KdV equation (Table \ref{tab:identification}). The large spikes in the loss curves during training (also studied by \cite{Byrd2016QuasiNewton}) indicate the transition from Adam to L-BFGS-B optimizers.}
    \label{fig:loss_kdv}
    \vspace{-3ex}
\end{figure}

\paragraph{Faster Convergence:}
To analyze the convergence speeds of both networks, Figure \ref{fig:loss_kdv} shows an example where the QRes network managed to reach comparable training loss as PINNs with roughly 1/5 of the training epochs as PINN on Adam optimizer. The QRes network reached final convergence on the L-BFGS-B optimizer with roughly 1/3 of the training epochs that PINNs used. We observed similar trends across other datasets, indicating the superior training convergence of QRes. 
\begin{table*}[ht!]
\begin{small}
\caption{\normalsize Comparing different network architectures for solving Burgers' equation.}
\begin{center}
\begin{tabular}{l|llll|lll}
\thickhline
Model       & PINN      & ISC       & QSC       & QRes-lite      & APINN     & QRes-full    & QRes-lite \\ \hline
Params      & 3.02k     & 3.02k     & 1.54k     & 1.54k     & 3.02k     & 2.94k     & 1.54k  \\ 
Optimizer   & L-BFGS-B  & L-BFGS-B  & L-BFGS-B  & L-BFGS-B  & Adam      & Adam      & Adam  \\ 
Error $u$   & 3.760e-03 & 1.504e-02 & 3.875e-03 & 3.386e-03 & 1.420e-01 & 8.630e-02 & 1.291e-01 \\ \thickhline
\end{tabular}
\label{tab:burgers-inference}
\end{center}
\vspace{-3ex}
\end{small}
\end{table*}

\begin{figure}[ht]
    \centering
    \includegraphics[width=0.49\textwidth]{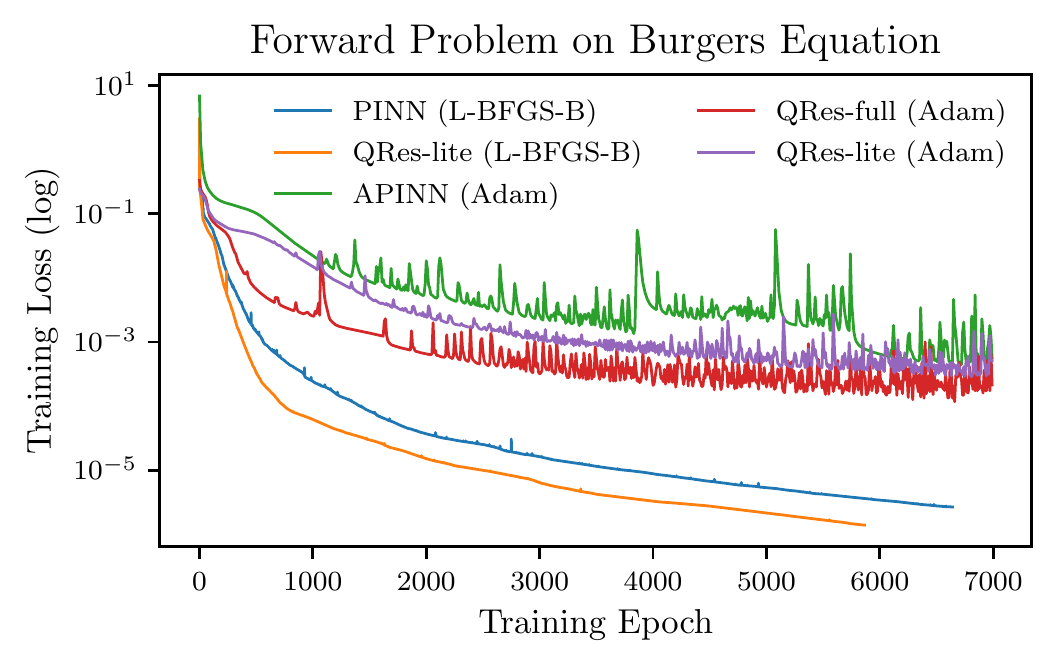}
    \vspace{-3ex}
    \caption{Training loss curves for solving Burgers' equation.}
    \label{fig:loss_burgers_inference}
    \vspace{-3ex}
\end{figure}

\paragraph{Comparison with Other Baselines:}
To demonstrate the advantages of QRes, we compared it with some other baselines for solving Burgers' equation. First, we considered APINN, which is PINN with \textit{adaptive activations}\cite{Jagtap_2020}
expressed as $H^{(l)} = \sigma[n~\alpha~(W H^{(l-1)} + b)]$, where $\alpha$ is the scaling parameter and $n$ is a hyperparameter. We followed the same settings of APINN as mentioned in the original work, i.e., $n=5$ and Adam optimizer.
Next, we used two baselines architectures that share similar ideas as ResNet \cite{he2016deep}. ISC, which is an abbreviation for \textit{Identity ShortCut}, has the closest resemblance to ResNet, as it adds the layer input to the activation, i.e., $H^{(l)} = \sigma[W H^{(l-1)} + b] + H^{(l-1)}$. QSC denotes \textit{Quadratic ShortCut}, which adds the quadratic residual after activation, i.e., $H^{(l)} = [W_1 H^{(l-1)}] \circ [W_2 H^{(l-1)}] + \sigma{[ W_1 H^{(l-1)} + b ]}$. 

For QRes networks, we tested two network sizes by adjusting network widths. The QRes-full has roughly the same number of parameters as the PINN and APINN, while QRes-lite has roughly half that number. Since the QSC also has two weight matrices, we set it to have the same width and depth as the QRes network (QRes-lite). While L-BFGS-B helped the models to reach higher accuracy, our experiments show that the APINN is very unstable when trained with L-BFGS-B. Therefore, we prepared QRes networks trained both with Adam and L-BFGS-B optimizers to have fair comparisons. We trained all models for 7k epochs with Adam, which is roughly the same number of epochs needed for L-BFGS-B to converge.

The results are shown in Table \ref{tab:burgers-inference} and Figure \ref{fig:loss_burgers_inference}. Both versions of the QRes networks outperform APINN when trained with Adam. For the group of models that were trained with L-BFGS-B optimizer, QRes-lite produces more accurate predictions than PINN and ISC, with smaller number of parameters. On the other hand, QSC performs even worse than PINN. 
In terms of convergence speed, Figure \ref{fig:loss_burgers_inference} further support that QRes networks are consistently faster than the baselines regardless of the choice of optimizers.

\paragraph{Analysis of Results:}

\begin{figure*}[ht]
    \begin{subfigure}[t]{0.25\textwidth}
        \centering
        \vspace{0.8cm}
        \label{fig:Navier_Stoke_exact_p}
        \includegraphics[width=1.0\textwidth]{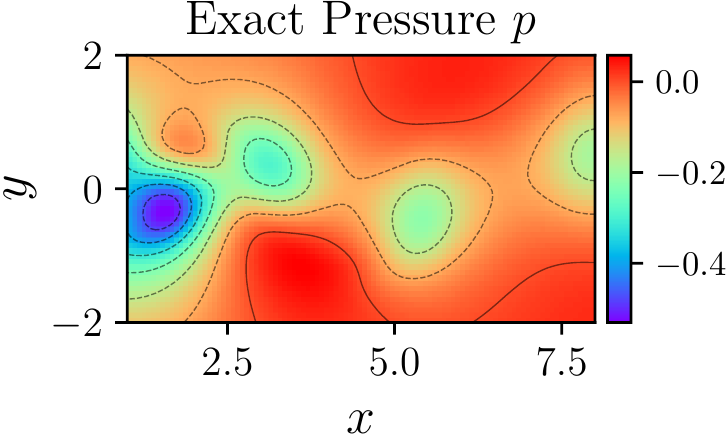}
    \end{subfigure}
    \ 
    \begin{subfigure}[t]{0.73\textwidth}
        \centering
        \label{fig:Navier_Stoke_error_map}
        \includegraphics[width=1.0\textwidth]{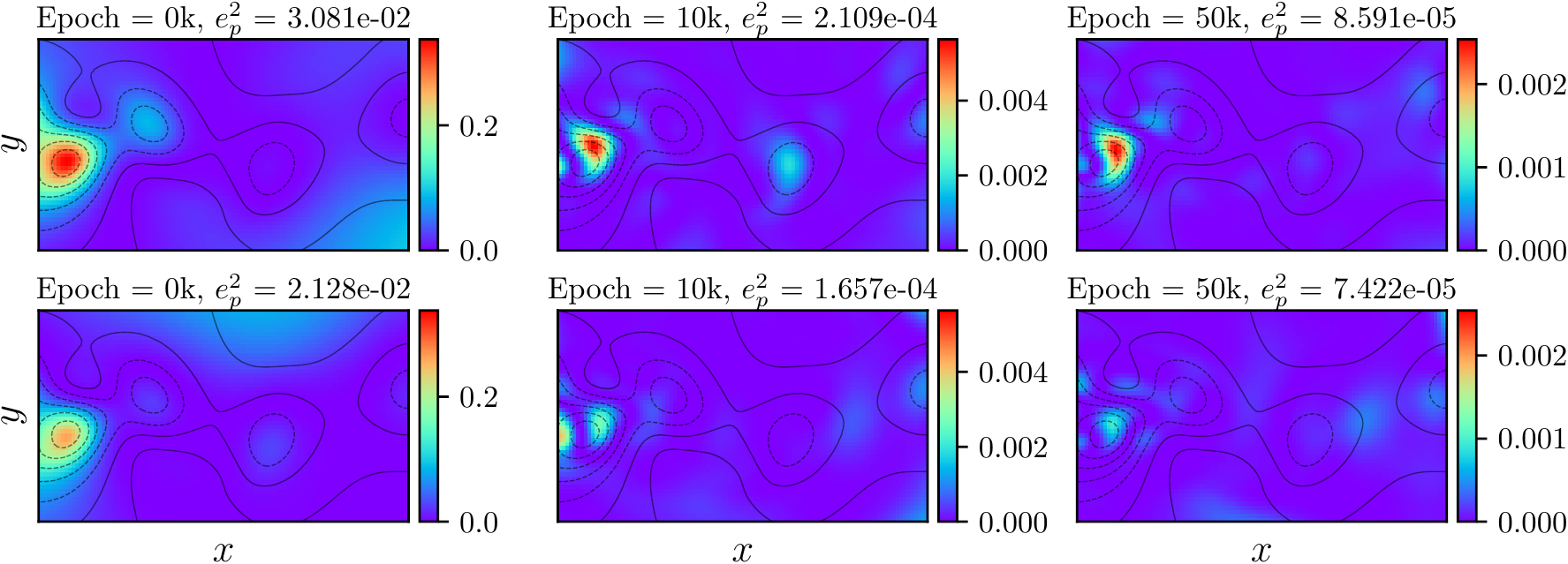}
    \end{subfigure}
    \caption{Visualizations of pressure field predictions for N-S equations. Left figure shows the exact pressure field while the remaining figures show error maps. The top row shows distribution of square error for PINN (1.38k parameters) and the bottom row is for the QRes networks (1.37k parameters). $e_p^2$ is the mean square error for the current epoch and dark contour lines represent the exact pressure field. Note that the scales of errors are different across epochs but are the same between QRes and PINN.} 
    \label{fig:error_map}
\end{figure*}

Figure \ref{fig:error_map} compares visualizations of pressure field predictions of PINN and QRes for N-S equations at different epochs in training. We can see the contour lines reveal a steep drop of pressure at the left of each figure (representing a region with high frequency patterns) where PINNs struggle to learn even after 50k epochs. On the other hand, QRes manages to digest the high frequency pattern (in regions where pressure values change abruptly) much faster than PINNs within 10k epochs. 

\begin{figure*}[!ht]
    \begin{subfigure}[b]{1.0\textwidth}
        \centering
        \includegraphics[width=1.0\textwidth]{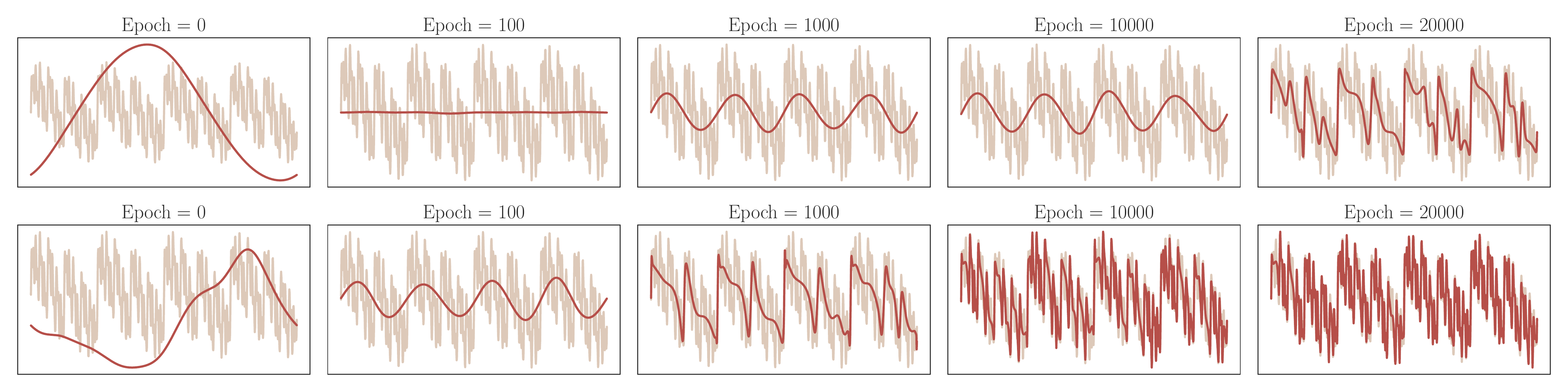}
        \caption{Predicted (dark) and ground truth (light) waves for DNN (top) and QRes (bottom).}
        \label{fig:sin_waves}
    \end{subfigure}
    \begin{subfigure}[b]{1.0\textwidth}
        \centering
        \includegraphics[width=1.0\textwidth]{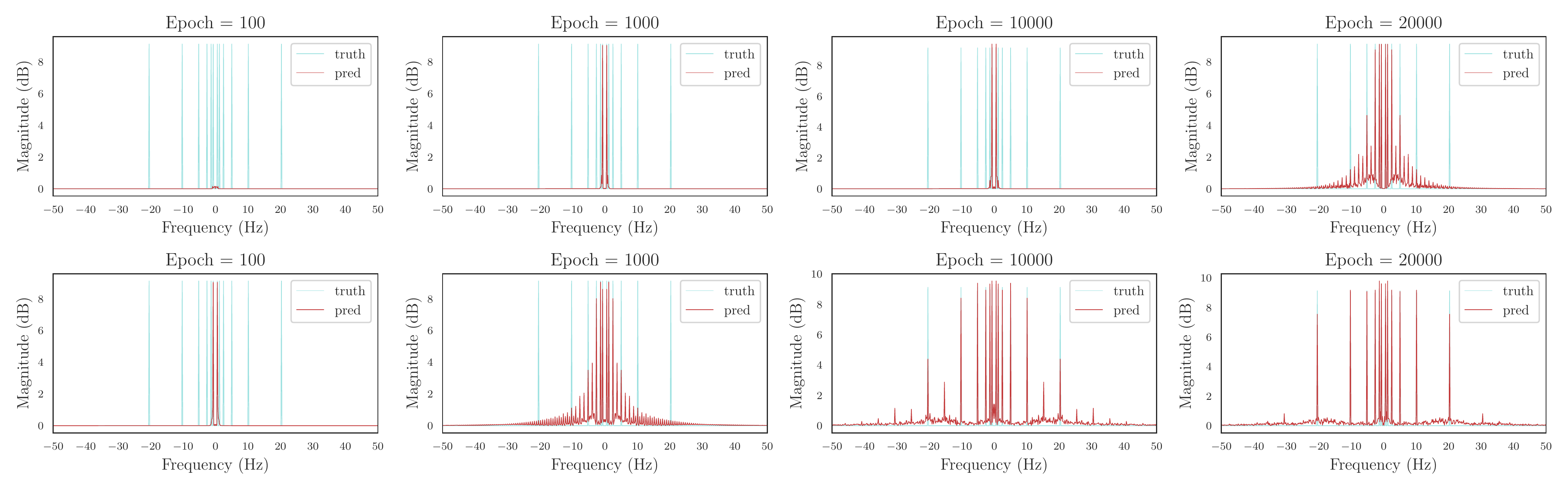}
        \caption{Spectrum of the predicted (red) and ground truth (blue) waves for DNN (top) and QRes (bottom). The spectrum is obtained using shifted double-sided fast Fourier transform (FFT) on 1000 points.}
        \label{fig:sin_waves_spectrum}
    \end{subfigure}
    \caption{Training a DNN (33.54k parameters) and a QRes network (33.21k parameters) to fit a superposition of sine waves with different frequencies: $4/2\pi, 8/2\pi, 16/2\pi, 32/2\pi, 64/2\pi$ and $128/2\pi$ Hz.}
    \label{fig:composition}
    \vspace{-1.5ex}
\end{figure*}

\subsection{Generalizing to General ML Problems:}
\label{sec:eval_patterns}

To analyze the ability of QRes and DNNs to learn higher frequencies in general ML problems, we performed a toy experiment to fit a composition of mono-frequency sine waves~\cite{convergence2019}. The task is to fit 1k data points on the composited curve (shown in \ref{fig:sin_waves}) using mean square error loss functions. All models are trained for 20k epochs with Adam optimizer. The results are shown in Figure \ref{fig:composition}, which shows two interesting characteristics of QRes. First, QRes learns higher frequencies much faster than DNNs, which supports our theoretical results. Second, like DNNs, the QRes exhibits the phenomena of \textit{spectral bias}\cite{convergence2019}, which is a well-known phenomenon that neural networks learn lower frequencies earlier in training, making it possible to apply early stopping to avoid overfitting. 

\section{Conclusion}
In this work, we proposed \textit{quadratic residual networks} (QRes) as a new type of neural networks with sufficiently high functional capacity (or expressive power). Using tools from algebraic geometry, we developed theories that prove the efficiency of the QRes networks. Following the original PINN work, we presented empirical evidences that QRes shows consistent advantages over deep neural networks in terms of parameter efficiency, convergence speed, and accuracy. Our work suggests that physics-informed deep learning can benefit from more nonlinearity in the network (as investigated by QRes), which can be the subject of future investigations. The remarkable advantage in learning higher frequencies further suggests the promise of using QRes networks in a broader range of ML applications. A full-length version of this article is available at \cite{jie2021qres}

\vspace{-1ex}
\section*{Acknowledgments}
This work was supported by NSF grant \#2026710.
\vspace{-3ex}

\bibliography{Jie, Anuj}
\bibliographystyle{ieeetr}

\section{Appendix}
\label{sec:appendix}
\subsection{Technical Proofs}
\label{sec:proofs}


Here we provide brief proofs for the theoretical results.

\textsc{\\Proposition \ref{pp:single_layer_QRes}}\ \ 
\textit{A single-layer linearly activated \textup{($r = 1$)} quadratic residual network of architecture $\bm{d}=(d_0, d_1)$ has a filling functional space of degree 2, i.e., its leading functional space $\mathcal{F}^{2}_{\bm{d}, 1} = \sym_{2}{(\mathbb{R}^{d_0})^{d_1}}$.}

\begin{proof} 
We can relate the linear QRes layer to a quadratic polynomial regression. Having linear indepence between $W_1$ and $W_2$, the functional space of a single layer QRes network has two subspaces of homogeneous polynomials: the linear term $W_1 H$ has $\sym_{1}{(\mathbb{R}^{d_0})^{d_1}}$ and the quadratic residual has $\sym_{2}{(\mathbb{R}^{d_0})^{d_1}}$, which is the leading functional space of the QRes network. Therefore, by \textit{Definition \ref{def:leading}}, it has a filling functional space of degree 2.
\end{proof}

\textsc{\\Lemma \ref{lm:efficiency}}\ 
\textit{For an activation function with leading degree $r \geq 1$ and network architecture $\bm{d} = (d_0, ..., d_h)$, the leading functional variety of a QRes network, $\mathcal{V}^{2}_{\bm{d}, r}$, and a neural network, $\mathcal{V}_{\bm{d}, r}$, satisfy $\mathcal{V}^{2}_{\bm{d}, r} = \mathcal{V}_{\bm{d}, 2r}$.}

\begin{proof}
This can be proven by discussing the equivalence of functional space for every layer in the network using Proposition \ref{pp:single_layer_QRes}. For the $i$-th layer in the QRes network, $i=1, 2, ..., h$, before applying nonlinear activation, it has $\mathcal{V}^{2}_{(d_{i-1}, {d_i}), 1} = \sym_{2}{(\mathbb{R}^{d_{i-1}})^{d_i}} = \mathcal{V}_{(d_{i-1}, d_i), 2}$ (since a single-layer neural network with polynomial activation of degree $2$ has a filling functional space of degree $2$). This proves the case for $r=1$. For nonlinear activations of leading degree $r$, applying the activation function to the space $\mathcal{V}^{2}_{(d_{i-1}, {d_i}), 1}$, we obtain: $\mathcal{V}^{2}_{(d_{i-1}, d_i), r} = \left(\mathcal{V}^2_{(d_{i-1}, d_i), 1}\right)^{\otimes r} = \left(\mathcal{V}_{(d_{i-1}, d_i), 2}\right)^{\otimes r} = \mathcal{V}_{(d_{i-1}, d_i), 2r}$, where $\otimes$ denotes Kronecker product. Since the relation applies to each layer, thus we have $\mathcal{V}^{2}_{\bm{d}, r} = \mathcal{V}_{\bm{d}, 2r}$.
\end{proof}

\textsc{\\Theorem \ref{th:depth_efficiency}}\ 
\textit{\textup{(Depth Efficiency)} For a fixed leading degree $r \geq 2$, let us assume that the functional variety of a quadratic residual network $\bm{d_q} = (d_0, .., d_{h_q})$ is filling, $\mathcal{V}^{2}_{\bm{d_q}, r}$ is the leading functional variety for the QRes network, and $\mathcal{V}_{\bm{d}, r}$ is the leading functional variety for a neural network $\bm{d_n} = (d_0', .., d_{h_n}')$, where $h_n, h_q > 1$ and $d_0 = d_0', d_{h_q} = d_{h_n}'$. If $\dim{\mathcal{V}_{\bm{d_n}, r}} \geq \dim{\mathcal{V}^{2}_{\bm{d_q}, r}}$, then $$h_n \geq 1 + \left( 1 + \frac{\log{2}}{\log{r}} \right)(h_q - 1)$$}
   
\begin{proof}
Let $d_h = d_{h_q} = d_{h_n}'$. Since the QRes network is filling for leading degree $r$, from Lemma \ref{lm:efficiency} we have $\dim{\mathcal{V}^{2}_{\bm{d_q}, r}} = \dim{\mathcal{V}_{\bm{d_q}, 2r}} = \dim{\sym_{(2r)^{h_q-1}}{(\mathbb{R}^{d_0})^{d_h}}}$. Meanwhile, the dimension of $\mathcal{V}_{\bm{d_n}, r}$ is at most that of its ambient output space, i.e., $\dim{\sym_{r^{h_n-1}}{(\mathbb{R}^{d_0})^{d_h}}}$. Thus,
\begin{align*}
    \dim{\mathcal{V}^{2}_{\bm{d_q}, r}} 
    &= \dim{\sym_{(2r)^{h_q-1}}{(\mathbb{R}^{d_0})^{d_h}}},\\
    &\leq \dim{\mathcal{V}_{\bm{d_n}, r}},\\
    &\leq \dim{\sym_{r^{h_n-1}}{(\mathbb{R}^{d_0})^{d_h}}}
\end{align*}
Thus, $(2r)^{h_q-1} \leq r^{h_n-1}$. Rearranging the terms yields the inequality in the theorem.
\end{proof}

\textsc{\\Proposition \ref{pp:bound_filling_with_qres}}\ \ 
\textit{\textup{(Minimal Filling Width)} For a neural network or a QRes network with architecture $\bm{d} = (d_0, ..., d_h)$ and leading activation degree $r \geq 2$, if 
$$
d_{h-i} \geq \min{\left[ d_h . r^{id_0}, \binom{r^{h-i} + d_0 - 1}{ r^{h-i} } \right]}    
$$
for each $i = 1, ..., h-1$, then its functional variety is filling, and we call the lower bound of $d_{h-i}$ as the minimal filling width at this layer with leading degree $r$.}

\begin{proof}
In Theorem 10 in \cite{kileel2019expressive}, it was proven that the above inequality holds for neural networks with polynomial activation degree $r$. This proof of minimal filling width is agnostic to the intermediate (hidden) layer structure (such as that in QRes networks), and only depends on the input and output dimensions $d_0, d_h$ and network depth $h$.
Since the quadratic residuals in QRes networks raise the input to second power without expanding the layer widths (Proposition \ref{pp:single_layer_QRes} \& Lemma \ref{lm:efficiency}) or increasing the degree of activation, the same proof applies to QRes networks with leading activation degree $r$.
\end{proof}



\begin{table*}[t]
\begin{center}
\caption{\normalsize Specifications of Experiments.}
\begin{small}
\begin{tabular}{l|l|l|l|l|l}
\thickhline
Source & Model & PDE / Problem & Optimizer  & Epochs  & Network Architecture\\ \hline
Fig. \ref{fig:pde_sin} &  NN   & Sine wave & Adam & 10k & (1, 20$_{\times 3}$, 1),  ELU\cite{clevert2016fast}   \\
Fig. \ref{fig:pde_sin} &  PINN & Sine wave & Adam & 10k & (1, 20$_{\times 3}$, 1), ELU\\ \hline
Tab. \ref{tab:identification} &  PINN & Navier-Stoke & Adam, L-BFGS-B & 200k & (3, 20$_{\times 8}$, 2), \textit{tanh}\\
Tab. \ref{tab:identification} &  QRes & Navier-Stoke & Adam, L-BFGS-B & 200k & (3, 14$_{\times 8}$, 2), \textit{tanh} \\
Tab. \ref{tab:identification} &  PINN & Burgers & Adam, L-BFGS-B & 10k & (2, 20$_{\times 8}$, 1),  \textit{tanh} \\
Tab. \ref{tab:identification} &  QRes & Burgers & Adam, L-BFGS-B & 10k & (1, 10$_{\times 8}$, 1),  \textit{tanh} \\
Tab. \ref{tab:identification} \& Fig. \ref{fig:loss_burgers_inference} &  PINN & Korteweg–de  Vries & Adam, L-BFGS-B & 50k & (1, 50$_{\times 4}$, 50), \textit{tanh} \\
Tab. \ref{tab:identification} \& Fig. \ref{fig:loss_burgers_inference} &  QRes & Korteweg–de  Vries & Adam, L-BFGS-B & 10k & (1, 20$_{\times 4}$, 50), \textit{tanh} \\ \hline
Tab. \ref{tab:inference} &  PINN & Schr\"{o}dinger & Adam, L-BFGS-B & 50k  & (2, 100$_{\times 4}$, 2), \textit{tanh} \\
Tab. \ref{tab:inference} &  QRes & Schr\"{o}dinger & Adam, L-BFGS-B & 50k  & (2, 50$_{\times 4}$, 2), \textit{tanh} \\
Tab. \ref{tab:inference} &  PINN & Allen-Cahn & Adam, L-BFGS-B & 10k  & (1, 200$_{\times 4}$, 101), \textit{tanh} \\
Tab. \ref{tab:inference} &  QRes & Allen-Cahn & Adam, L-BFGS-B & 10k  & (1, 50$_{\times 4}$, 101), \textit{tanh} \\ \hline
Tab. \ref{tab:burgers-inference} \& Fig. \ref{fig:loss_burgers_inference} &  PINN & Burgers & L-BFGS-B & ftol  & (2, 20$_{\times 8}$, 1), \textit{tanh} \\
Tab. \ref{tab:burgers-inference} \& Fig. \ref{fig:loss_burgers_inference} &  QRes-lite & Burgers & L-BFGS-B & ftol  & (2, 10$_{\times 8}$, 1), \textit{tanh} \\
Tab. \ref{tab:burgers-inference} \& Fig. \ref{fig:loss_burgers_inference} &  ISC & Burgers & L-BFGS-B & ftol  & (2, 20$_{\times 8}$, 1), \textit{tanh} \\
Tab. \ref{tab:burgers-inference} \& Fig. \ref{fig:loss_burgers_inference} &  QSC & Burgers & L-BFGS-B & ftol  & (2, 10$_{\times 8}$, 1), \textit{tanh} \\
Tab. \ref{tab:burgers-inference} \& Fig. \ref{fig:loss_burgers_inference} &  APINN & Burgers & Adam & 7k  & (2, 20$_{\times 8}$, 1), \textit{tanh} \\
Tab. \ref{tab:burgers-inference} \& Fig. \ref{fig:loss_burgers_inference} &  QRes-full & Burgers & Adam & 7k  & (2, 14$_{\times 8}$, 1), \textit{tanh} \\
Tab. \ref{tab:burgers-inference} \& Fig. \ref{fig:loss_burgers_inference} &  QRes-lite & Burgers & Adam & 7k  & (2, 10$_{\times 8}$, 1), \textit{tanh} \\ \hline
Fig. \ref{fig:burgers_efficiency} &  PINN   & Burgers & L-BFGS-B & ftol  & Variable,  \textit{tanh}   \\
Fig. \ref{fig:burgers_efficiency} &  QRes & Burgers & L-BFGS-B & ftol  & Variable, \textit{tanh}\\ \hline
Fig. \ref{fig:error_map} &  PINN & Navier-Stoke & Adam & 50k & (3, 20$_{\times 4}$, 2),  \textit{tanh}   \\
Fig. \ref{fig:error_map} &  QRes & Navier-Stoke & Adam & 50k & (3, 14$_{\times 4}$, 2), \textit{tanh}\\ \hline
Fig. \ref{fig:composition} &  NN   & Composited wave & Adam & 20k & (1, 128$_{\times 3}$, 1),  \textit{tanh}   \\
Fig. \ref{fig:composition} &  QRes & Composited wave & Adam & 20k & (1, 90$_{\times 3}$, 1), \textit{tanh}\\
\thickhline      
\end{tabular}
\label{tab:specifications}
\end{small}
\end{center}
\end{table*}

\textsc{\\Theorem \ref{th:width_efficiency}}\ 
\textit{\textup{(Width Efficiency)} Suppose a neural network $\bm{d_n} = (d_0, .., d_{h})$ is filling for leading activation degree $r \geq 2$. Given a quadratic residual network $\bm{d_q} = (d_0', .., d_{h}')$ with $d_0 = d_0'$ and  $d_{h} = d_{h}'$, such that $\dim{\mathcal{V}_{\bm{d_n}, r}} = \dim{\mathcal{V}^{2}_{\bm{d_q}, r}}$. Suppose $\bm{d_n}$ is a minimal filling architecture, then for each $i = 1, ..., h-1$,  
$$
\lim_{r \to \infty}{d_{h-i}} = O(2^{\tau}) \lim_{r \to \infty}{d_{h-i}'}    
$$
where $\tau = \min{\left[ id_0, (h-i)(d_0-1) \right]}$.}

\begin{proof}
To understand the limiting behavior at large values of $r$  ($r \to \infty$), note that $d_h r^{id_0} = O(r^{id_0})$ since $d_0, d_h$ are constants. Further, 
\begin{align*}
    \binom{r^{h-i} + d_0 - 1}{ r^{h-i}} &= \frac{(r^{h-i} + d_0 - 1)!}{(r^{h-i})!(d_0 - 1)!}\\
                                        &= \frac{\prod_{k=1}^{d_0 - 1}{(r^{h-i} + k)}}{(d_0 - 1)!} \\
                                        &= O(r^{(h-i)(d_0 - 1)})
\end{align*}
Then, by Proposition \ref{pp:bound_filling_with_qres}, we get the following inequality:
\begin{align*}
    d_{h-i} &\geq \min{\left[ d_h r^{id_0}, \binom{r^{h-i} + d_0 - 1}{ r^{h-i} } \right]}  = O(r^\tau),   
\end{align*}
where $\tau = \min{[id_0, (h-i)(d_0 - 1)]}$.

Since the neural network is filling, we can state that $\mathcal{V}_{\bm{d_n}, r} = \dim{\sym_{r^{h-1}}{(\mathbb{R}^{d_0})^{d_h}}}$.
With the same depth, if the QRes network is also filling, then we will get $\mathcal{V}^{2}_{\bm{d_q}, r} = \sym_{(2r)^{h-1}}{(\mathbb{R}^{d_0})^{d_h}}$, which contradicts with the given statement that $\dim{\mathcal{V}_{\bm{d_n}, r}} = \dim{\mathcal{V}^{2}_{\bm{d_q}, r}}$. This implies that $\bm{d_q}$ is not filling for $r$. Let us suppose the QRes network is filling for a lower degree $r' < r$, and $r'$ is the highest degree of ambient space that $\bm{d_q}$ can fill, i.e., $\bm{d_q}$ is a minimal filling architecture for degree $r'$. From the given statement $\dim{\mathcal{V}_{\bm{d_n}, r}} = \dim{\mathcal{V}^{2}_{\bm{d_q}, r}}$, we then have 
\begin{equation*}
    \dim{\sym_{r^{h-1}}{(\mathbb{R}^{d_0})^{d_h}}} = \dim{\sym_{(2r')^{h-1}}{(\mathbb{R}^{d_0})^{d_h}}}.
\end{equation*}
Thus, $r' = r/2$. Since $\bm{d_n}$ is a minimal filling architecture, we have $d_{h-i} = O(r^\tau) = O(2^\tau) O(r'^\tau)$, which is equal to $O(2^\tau) d_{h-i}'$ for large values of $r$.
\end{proof}


\subsection{Additional Specifications of Experiments}

For reproducibility of the results reported in the paper, we provide additional specifications of our experiments in Table \ref{tab:specifications}. We also make the following additional remarks.

\begin{itemize}
    \item Table \ref{tab:problems} describes the PDE problems we empirically studied in this work. Refer to the original PINN work\cite{raissi2019physics} and its corresponding Github page\footnote{\url{https://maziarraissi.github.io/PINNs/}} for more information on the details of these PDEs.
    \item Learning rates are set to be default values ($0.001$) for Adam optimizer in all our experiments. 
    \item The termination condition \textit{ftol} for L-BFGS-B is when the loss values barely change with parameter updates, i.e., 
    $$\frac{(L_k - L_{k+1})}{\max{(|L_k|,|L_{k+1}|,1)}} \leq \textup{ftol}.$$
    where $L_k$ is the loss value at $k$-{th} epoch. We use $\textup{ftol} \approx 2.22 \times 10^{-16}$.
\end{itemize}

\subsection{Remarks on Computational Time}

Based on our experiments on NVIDIA TITAN RTX GPU, with less number of parameters, the QRes networks are faster than deep neural networks during test time (despite calculating Hadamard product), which is crucial for deployment on devices with limited computational power. For the training phase, due to more complicated computational graph for gradient back-propagation, the running time per epoch needed for QRes is typically 1.5 $\sim$ 3.0 times that of DNN with roughly the same number of parameters. However, since the QRes networks converge much faster than NNs (as is shown in Figure \ref{fig:loss_burgers_inference}), we can terminate the training earlier. Hence, the QRes networks often need less training time than DNNs.
\newpage

\end{document}